\documentclass[11pt,twoside,twocolumn,a4paper]{article}

\usepackage{cvww_arxiv}
\usepackage{times}
\usepackage{color}
\usepackage{epsfig}
\usepackage{graphicx}
\usepackage{amsmath}
\usepackage{amssymb}
\usepackage{float}
\usepackage{animate}
\usepackage{subfig}
\usepackage{textcomp}
\usepackage{url}

\makeatletter
\renewcommand{\paragraph}{%
  \@startsection{paragraph}{4}%
  {\z@}{0.5ex \@plus 1ex \@minus .2ex}{-1em}%
  {\normalfont\normalsize\bfseries}%
}
\makeatother

\newcommand{\figref}[1]{Figure~\ref{#1}}
\newcommand{\tblref}[1]{Table~\ref{#1}}
\newcommand{\sref}[1]{Section~\ref{#1}}

\usepackage[pagebackref=true,breaklinks=true,bookmarks=false]{hyperref}

\cvwwfinalcopy

\ifcvwwfinal\fi

\begin{document}

%%%%%%%%% TITLE
\title{Camera-based vehicle velocity estimation from monocular video}

\author{Moritz Kampelm\"uhler$^1$ \hspace{30pt} Michael G. M\"uller$^2$ \hspace{30pt} Christoph Feichtenhofer$^1$\vspace{5pt}\\ 
$^1$Institute of Electrical Measurement and Measurement Signal Processing\\
$^2$Institute of Theoretical Computer Science\\
Graz University of Technology\\
{\tt\small kampelmuehler@student.tugraz.at},\hspace{5pt}{\tt\small mueller@igi.tugraz.at},\hspace{5pt}{\tt\small feichtenhofer@tugraz.at}\\
}

\maketitle
\ifcvwwfinal\fi

%%%%%%%%% ABSTRACT
\begin{abstract}
	This paper documents the winning entry at the CVPR2017 vehicle velocity estimation challenge.\\
   Velocity estimation is an emerging task in autonomous driving which has not yet been thoroughly explored. The goal is to estimate the relative velocity of a specific vehicle from a sequence of images.
	 In this paper, we present a light-weight approach for directly regressing vehicle velocities from their trajectories using a multilayer perceptron.  Another contribution is an explorative study of features for monocular vehicle velocity estimation. We find that light-weight trajectory based features outperform depth and motion cues extracted from deep ConvNets, especially for far-distance predictions where current disparity and optical flow estimators are challenged significantly. Our light-weight approach is real-time capable on a single CPU and outperforms all competing entries in the velocity estimation challenge. On the test set, we report an average error of 1.12 m/s which is comparable to a (ground-truth) system that combines LiDAR and radar techniques to achieve an error of around 0.71 m/s.
\end{abstract}

%%%%%%%%% BODY TEXT

\section{Introduction}
Camera sensors provide an inexpensive yet powerful alternative to range sensors based on LiDAR or radar. While LiDAR systems can provide very accurate measurements, they may also malfunction under adverse environmental conditions such as fog, snow, rain or even exhaust gas fumes \cite{rasshofer11, hasirlioglu17}. Arguably vision based sensing is more closely related to how humans engage in driving situations and it should thus be possible to solve any task in autonomous driving based on visual input.

\begin{figure}[t!]
	\centering
	\includegraphics[width=0.49\textwidth]{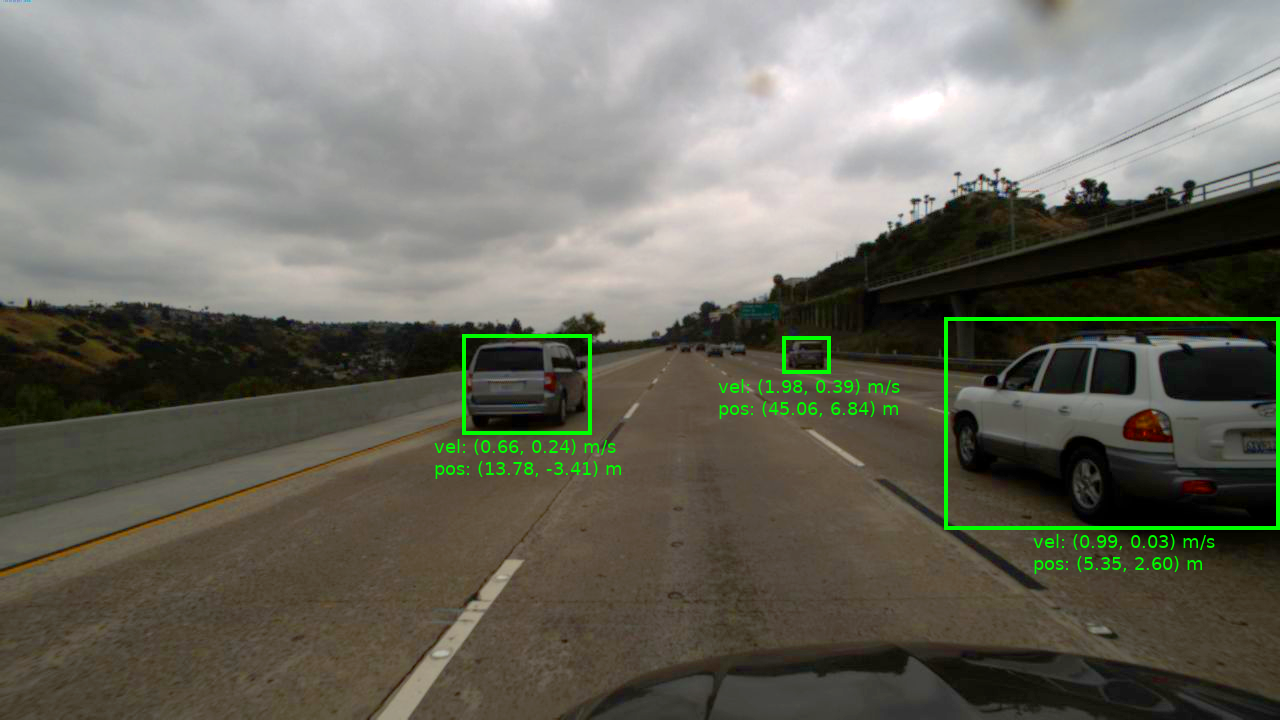}
	\vspace{-20pt}
	\caption{A sample image from a training sequence. Velocity and position ground truth are provided for the vehicles surrounded by the green bounding boxes.}
	\label{fig:sample}
	\vspace{-15pt}
\end{figure}

This work addresses monocular vehicle velocity estimation, an emerging task in autonomous driving which has not yet been thoroughly explored. The specific task, which forms the base for this work, was introduced as the Autonomous Driving Velocity Estimation Challenge\footnotemark~at CVPR2017. The goal is to estimate the relative velocity of a specific vehicle from a sequence of monocular RGB images to aid autonomous driving algorithms such as for example collision avoidance \cite{aufrere03} or adaptive cruise control \cite{jurgen06}. \figref{fig:sample} shows an example image from the data\footnotemark[\value{footnote}].
\footnotetext{\url{http://benchmark.tusimple.ai/\#/t/2}}

Vehicle velocity estimation as such is not a new subject of interest, since it is extensively studied in the context of traffic surveillance \cite{hsieh06, coifman98}, where , however, a stationary camera is employed. Under the restriction of a fixed camera pose the problem becomes significantly less complex, since with a calibrated camera system angular measurements can be obtained and from these measurements velocity estimates can readily be established. In contrast in our case the observer resides on a moving platform and inferring velocity in a similar fashion would require additional information such as camera pose, ego-motion and foreground-background segmentation. Very recent research \cite{zhou17} shows that  estimating ego-motion as well as disparity maps from monocular camera images by means of \textit{structure from motion} is indeed possible, but still limited. Semantic segmentation of scenes, which is a fundamental problem in computer vision, has also more recently been tackled using deep neural networks \cite{dai16, li16}.

In a more general sense the given task can be seen as a lightweight version of \textit{object scene flow} as for example provided in the KITTI benchmark \cite{KITTI12, menze15}. Object scene flow aims at estimating dense 3D motion fields, which in their temporal evolution carry highly valuable information about the geometric constellation of a given scene. Recent approaches \cite{vogel13, vogel14} yield impressive results, but they rely on the availability of stereo image data. Furthermore, they come at the price of very high computational cost, such that the estimation for a temporal frame pair might take 5-10 minutes on a single CPU core. In autonomous driving scenarios computational resources are in general highly limited \cite{grzywaczewski17}, which makes object scene flow currently not practically feasible.

In this work we adopt recent deep learning architectures \cite{flownet17,monodepth17} for depth and motion estimation to leverage a mapping of the video input into a beneficial feature space for learning from the few training samples provided. Our approach employs a two-stage process for monocular velocity estimation. In a first step we extract vehicle tracks as well as dense depth and optical flow information, followed by locally aggregating these depth and motion cues at the tracked vehicle locations and concatenating over the temporal dimension. After this feature extraction procedure we use the spatiotemporal depth, flow and location features to train a fully connected regression network for velocity estimation of the respective vehicles.

Further on we conduct an extensive ablation study to investigate the impact of the individual features and combinations thereof on the regression performance as well as on the runtime of the estimation. We show that a light weight implementation can achieve excellent results, and that leveraging deep motion and depth cues does not necessarily improve performance for this task on the given data.

\section{Related Work}

\begin{figure*}[t]
	\centering
	\includegraphics[width=0.95\textwidth]{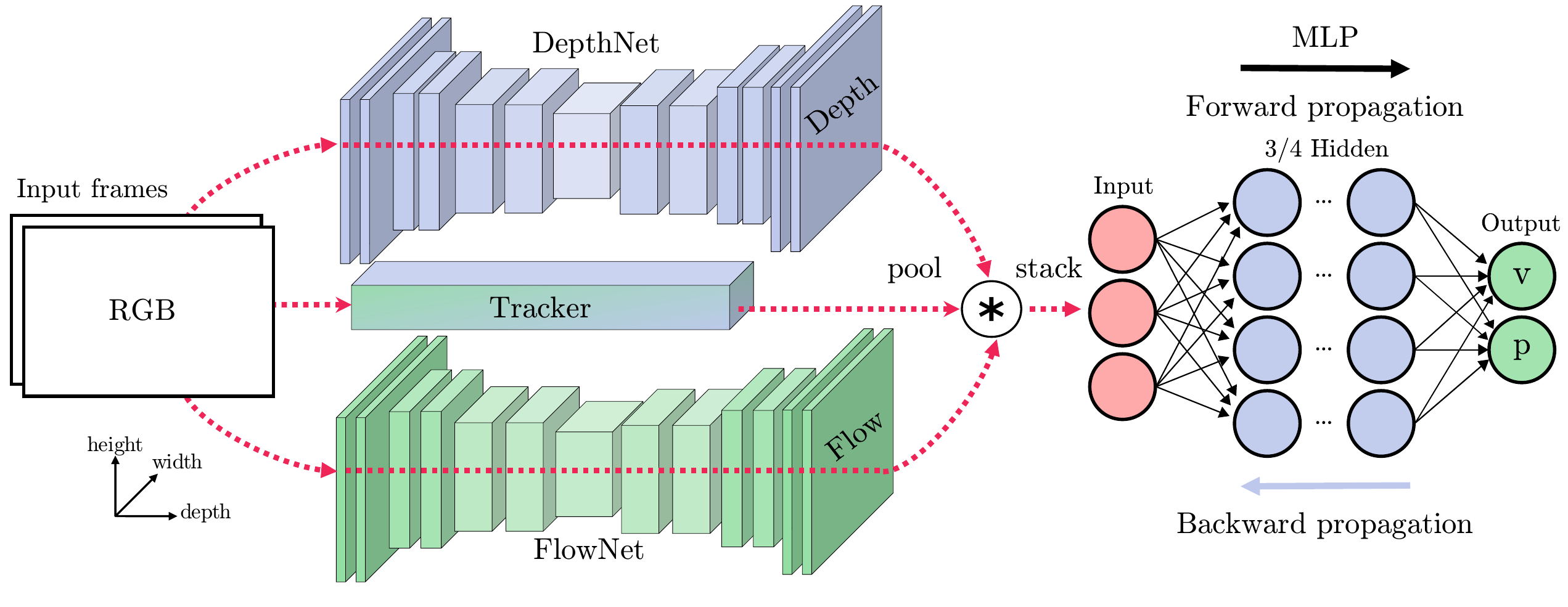}
	\vspace{-12pt}
	\caption{Overview of our overall architecture (see \sref{sec:approach} for details). Our proposed light-weight architecture only uses the trajectory features estimated by the tracker. }
	\vspace{-10pt}
	\label{fig:approach}
\end{figure*}

\paragraph{Tracking.}

Object tracking is one of the fundamental problems in computer vision and has been extensively studied \cite{yilmaz06} and applied in many different tasks.

\emph{Median Flow} \cite{medianflow10} is a method building on top of the \textit{Lucas-Kanade} \cite{Lucas81} method, which is an early optical flow algorithm operating on local intensity changes. This method is extended by a Forward-Backward error, which denotes the deviation between the trajectories obtained by tracking from $I_{t-1} \rightarrow I_t$ and $I_t \rightarrow I_{t-1}$. Using Forward-Backward error, robust predictions can be identified.

The \emph{Multiple Instance Learning} tracker \cite{MIL09} first transforms the image into an appropriate feature space, and uses a classifier as well as a motion model to determine the presence of an object in a frame, which is referred to as tracking by detection.

More recent methods like \cite{held16} employ convolutional neural networks to learn motion and appearance of objects. The feature maps of higher convolutional layers provide robust and accurate appearance representations, but lack spatial resolution. Lower layers on the other hand provide higher spatial resolution and less refined appearance representations. This hierarchical structure is used in \cite{ma15} by inferring responses of correlation filters on each corresponding layer pair.

\paragraph{Monocular Depth Estimation.}

Estimating the depth information of a scene seen from a given angle using only a single camera is not a well-posed problem. Some of the methods tackling this problem \cite{saxena06, xu17} apply supervised learning regimes requiring ground truth depth data. Since acquiring such ground truth data with sufficient accuracy requires immense effort, several methods are being developed that require little to no supervision.

In \cite{kuznietsov17} a semi-supervised approach is described that provides a fusion between using sparse ground truth data from a LiDAR sensor and stereo view synthesis, \ie estimating one image in a stereo pair from the other, to infer dense depth maps. Others \cite{monodepth17, xie16} in turn rely solely on stereo as a supervision signal, which comes with the benefit of easily available or obtainable data.

Some recent work \cite{zhou17, garg16} shows the capability of inferring single image depth from monocular video only. This is achieved by leveraging the (small) temporal motion of the camera and its thus changing pose to learn from multiple views of the scene. Via novel view synthesis the future camera frames can be used as a self-supervision signal. The mapping thus learned implicitly carries information about the 3D scene geometry.

\paragraph{Optical Flow Estimation.}

Optical flow is widely used in computer vision, \eg for video object detection \cite{zhu17}, to quantify pixel-wise motion in between frames of a moving scene. Traditional methods \cite{horn81,wedel09} employ variational motion estimation approaches, that regard the estimation of pixel level correspondences between frames as an optimization problem. 

With the growing interest in deep learning also optical flow estimation is now often successfully treated as a supervised learning problem \cite{dosovitskiy15}. This is achieved by employing convolutional neural networks for feature extraction and aggregation, followed by an 'upconvolutional' network, which concatenates the feature maps from the corresponding convolutional layers and jointly applies transposed convolution to increase spatial resolution. Further improvements on this approach have since been made \cite{flownet17} that provide improved performance and robustness as well as scalability.

\section{Vehicle velocity estimation} \label{sec:approach}

We present an explorative study over features for vehicle velocity estimation from monocular camera videos and discuss effectiveness as well as significance of the methods employed. The task is to estimate the relative velocity as well as position of given vehicles seen in short dashcam video snippets. Our overall approach, shown in \figref{fig:approach}, consists of a two stage process: First, features subsidiary to the task are extracted to subserve the second stage, which is a light-weight Multilayer Perceptron (MLP) architecture working on these features to regress velocity and positions of vehicle instances.

\subsection{Feature Extraction}

For the estimation of vehicle velocity, the raw RGB video data is first transformed into a beneficial feature space. We use three feature types of complementary nature: Vehicle tracks (\ie trajectories of the 2D object outline over time), depth (\ie disparity estimates from monocular imagery) and motion (\ie optical flow estimates between consecutive frames). The remainder of this section describes the specific algorithmic instances used for extracting these cues.

\paragraph{Tracks.}

For a given vehicle defined by a bounding box in a single frame, tracking over the temporal extent of the input serves for all further processing steps. A variety of well functioning tracking algorithms are readily available in literature. Since we aim for a lightweight tracker that should precisely localize the object outline, we employ fast trackers that operate directly at the pixel level, the \emph{Median Flow} \cite{medianflow10} and \emph{MIL} \cite{MIL09} trackers, both implemented in the \texttt{OpenCV} library \cite{opencv00}. 
The \emph{Median Flow} tracker comes with the benefit of being able to adapt bounding boxes over the trajectories, and provides a tight bounding box that can be a very useful feature when estimating the relative velocity of the objects. However, this tracker is unstable for occlusions since it employs forward-backward tracking. Whenever \emph{Median Flow} detects a tracking failure the missing bounding boxes are substituted with \emph{MIL} tracks. 

\paragraph{Depth.}

For dense depth map prediction we employ a recently described deep architecture \cite{monodepth17}, that learns monocular depth map prediction via novel view synthesis in a stereo environment. This is achieved by synthesizing \eg the left camera image from the right, where the left camera image is used as supervision signal. The warping operation that is thus learned implicitly carries information on the disparities in the source image. We use a model pre-trained on \emph{KITTI} \cite{KITTI12} and \emph{Cityscapes} \cite{cityscapes16} stereo images for predicting the dense disparity maps (note that this model is trained without disparity ground-truth). Owing to architectural constraints and limited computational capacity the RGB input images are resized to 512x256p. 

\begin{figure}[t]
\centering
\includegraphics[width=0.49\textwidth]{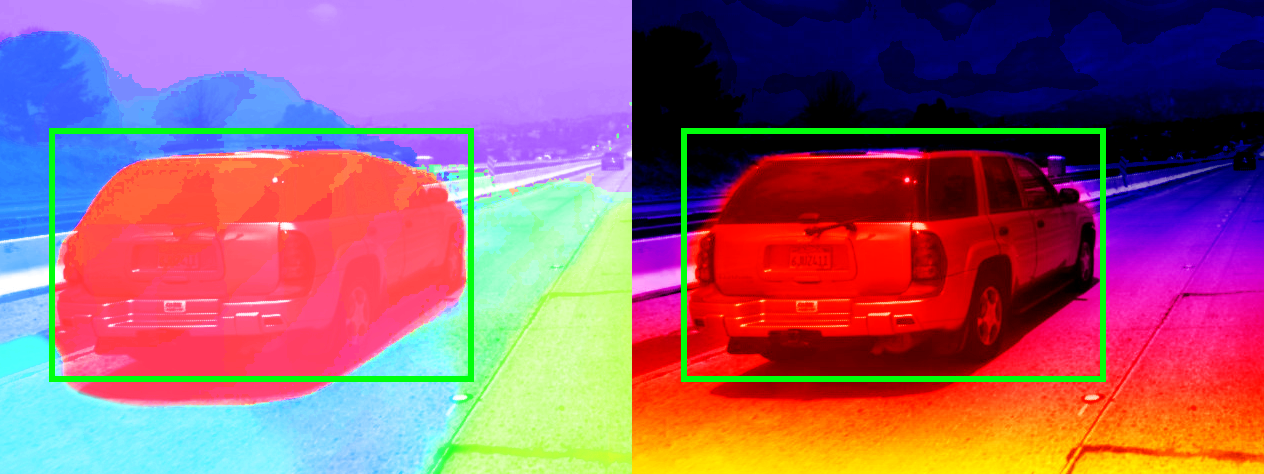}
\vspace{-20pt}
\caption{Optical flow (left, Middlebury flow encoding), bounding box and depth map (right, darker values represent larger distances) of a close range sample vehicle overlayed over the RGB image.}
\label{fig:features}
\vspace{-10pt}
\end{figure}

\paragraph{Motion.}

Finally, we extract motion information by extracting dense optical flow maps using a state-of-the-art neural network architecture, \emph{FlowNet2} \cite{flownet17}. \emph{FlowNet2} treats optical flow estimation as a supervised learning problem, where a convolutional neural network is trained on a volume of two stacked input frames with ground truth optical flow as a supervision signal.
In our case we use a \emph{FlowNet2} architecture pre-trained on the synthetic \emph{Flying Chairs} \cite{dfib15} dataset to calculate 39 dense $u,v$ flow maps from 512x256p input images.
A sample of the extracted feature maps is shown in \figref{fig:features}. In both feature maps the vehicle can be segmented from the background, but the capabilities are limited to close ranges.

\paragraph{Transformation into feature space.}

The depth and motion cues are  computed globally and further processed to serve as lightweight input features for a regression model. This is achieved by locally aggregating the dense predictions within the bounding boxes established by the tracking stage. The local aggregation is achieved by calculating the mean over the estimates within each tracked bounding box, after shrinking the box by 10\% in width and height, which reduces the variance since flow and depth cues tend to be inaccurate at the object boundaries. Subsequently, the aggregated feature vectors are temporally smoothed using a Gaussian kernel of width $5$, which is chosen in correspondence to the frame skip of $5$ in the learning stage. For optical flow, this procedure is carried out individually for each the horizontal and vertical component $u,v$. The temporal smoothing provides robustness to short-term deviations of the camera orientation (\eg caused by bumps when driving on a highway).

\begin{table*}[t]
\vspace{-5pt}
\centering
	\begin{tabular}{l|c|c|c|c|c|c|c}

	Features & (a) & (b) & (c) & (d) & (e) & (f) & (g) \\ 
	\hline 
	Tracking     & \checkmark &  &  & • & \checkmark  & \checkmark  & \checkmark  \\ 

	Motion & •          & \checkmark & •          & \checkmark & \checkmark  & • & \checkmark  \\ 
	Depth        & •          & •          & \checkmark & \checkmark  & & \checkmark  & \checkmark  \\ 
	%\hline 
	%Activation     & • & • & • & • & • & • & • \\ 
	\hline 
	$E_{V}$        & 1.86 & 4.68 & 4.59 & 3.74 & 1.90 & \textbf{1.83} & 1.86 \\ 
	\hline
	$E_{V,near}$   & 0.93 & 1.42 & 0.87 & \textbf{0.66} & 0.91 & 0.70 & 0.77 \\ 
	\hline
	$E_{V,medium}$ & \textbf{0.76} & 4.84 & 4.88 & 4.29 & 0.87 & 0.86 & 0.90 \\ 
	\hline
	$E_{V,far}$    & 3.43 & 6.87 & 7.72 & 6.20 & 3.24 & 3.23 & \textbf{3.06} \\ 
	\hline 

	\end{tabular}
	\vspace{-7pt}
	\caption{\label{tab:best_results} Best results achieved in ablation study for each individual range.}
	\vspace{-12pt}
\end{table*}

\subsection{Neural Network Model}

These pre-computed features allow us to use relatively shallow
fully-connected neural networks. This is especially advantageous for the task at hand, in
which the number of given training examples is relatively small and learning can take advantage of highly abstract features. We employ a comparatively simple, rather small and thus efficient 4-layer MLP architecture to regress from feature space to vehicle velocities.

\begin{figure}[t!]
\begin{center}
\includegraphics[width=0.45\textwidth]{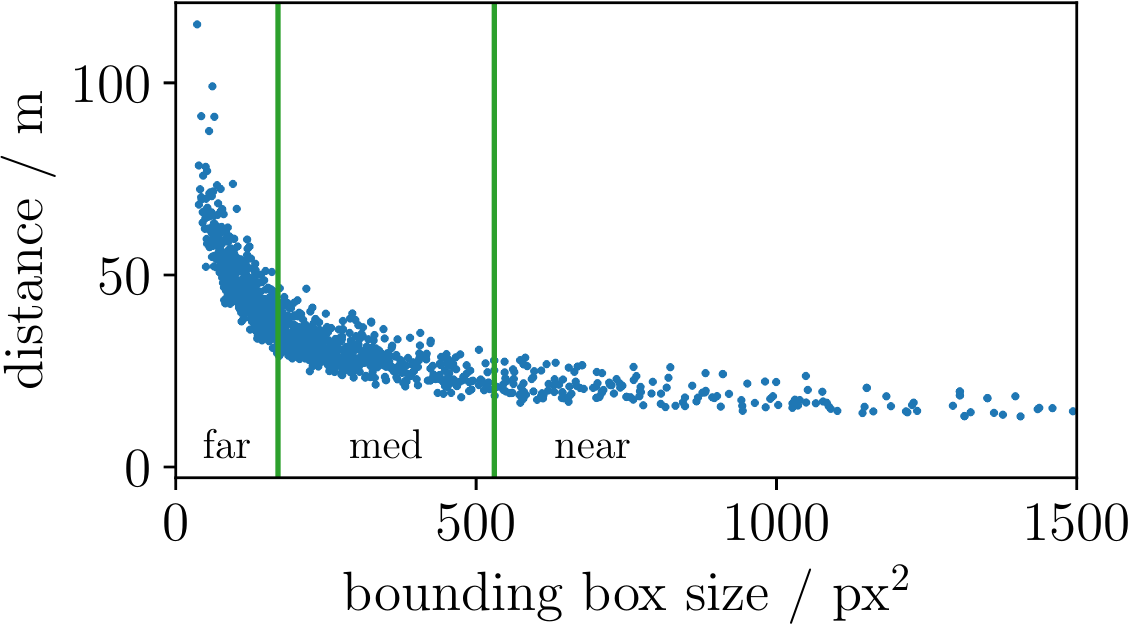}
\caption{Split of distance categories for neural network. For the labeled
training set, the true distance is known. The empirically chosen boundaries are
shown in green.}
\vspace{-10pt}
\label{fig:nmf_split}
\end{center}
\end{figure}

\paragraph{Data Split.} The relationship of the pre-computed features to the
learning output is highly nonlinear. Consider for instance the size of bounding
boxes around other vehicles: they rapidly decrease when the vehicle is close to
the camera, while remaining more or less constant if the vehicle is far away
regardless of the velocity. To facilitate learning in these distinct regimes,
we split the data set into three disjoint parts, based on the distance from the observer (near/mid/far), and train separate models for each. \figref{fig:nmf_split} shows the bounding box size as a function of the distance from the observer for the training set, as well as our chosen splits.
As indicator for the relative distance of the target vehicle we use the bounding box area of the last frame that the vehicle is shown. Although there is some variance near the borders between different sets, this split simplifies the learning task compared to a single model that has to work across all distances. % We will experimentally verify that in \sref{sec_experiments}. \todo{not really}

\paragraph{Architecture.} On each of these three sets, we performed
cross-validation to obtain a good network architecture. The resulting network
topologies are (number of hidden layers $\times$ number of units per layer)
$3\times40$ (near), $4\times60$ (medium), and $4\times70$ (far). Extensive
validation experiments revealed that best results could be obtained with
concatenated rectified linear units (CReLU) \cite{shang2016understanding} as hidden
unit activations; thus, all layer output sizes are twice the input size. The output layer
activation is linear. We use an L2 loss to train the network by using the ground-truth velocity values, obtained from a combination of LiDAR and radar sensors. 

\paragraph{Training.} Each individual network is trained on the MSE between network output and targets for 2000 epochs using minibatches of 50 samples. For regularization, weight decay of $10^{-5}$ and Dropout \cite{srivastava2014dropout} of 0.2 were
used. We train using the ADAM optimizer \cite{kingma2014adam}.

To make use of all available training data, we use a partitioning scheme similar
to k-fold cross-validation and train multiple networks per distance set. Each of
them is split up into five partitions. Four of these are used as training set,
the fifth is used for validation. After training for 2000 epochs on each
possible combination, the model with the lowest validation error is saved. This
results in $3\times5$ models for the entire dataset. Note that the number of
examples per neural network is quite small, so overfitting may occur easily as
the validation score can no longer be regarded as an accurate estimate of the
error on the test set in most cases.

The training data not only contains targets for the target vehicle velocity,
but also for the relative vehicle position. In other machine learning domains,
it is well-known that auxilliary targets can increase the learning performance
(see \eg for reinforcement learning
\cite{mirowski2016learning,jaderberg2016reinforcement,dosovitskiy2016learning}). We also make use
of all targets during training and simultaneously regress for vehicle velocities and distance. Early stopping is performed using only the
performance on the actual targets. This leads to slight increases in
performance.

\paragraph{Model Averaging.} Above, we described how $3\times5$ partitions are
trained. When evaluating the test set, we split the data according to the
computed bounding box areas using the same procedure as for the training set.
Then, the average over all five models for the respective distance is computed.
This gives the final estimation for the relative vehicle velocity.

\section{Experiments} \label{sec_experiments}

\subsection{Dataset} \label{sec:dataset}

The provided dataset includes 1074 driving sequences in freeway traffic, recorded by a single HD (1280x720p) camera, each 40 frames long captured at 20fps, as well as camera calibration matrices. For each sequence specific vehicles are annotated with a bounding box as well as position and velocity in both $x$ and $y$ coordinates in the last frame only, resulting in a total of 1442 annotated vehicles. For evaluation the individual vehicles are classified into three clusters according to their ground truth relative distance $d$ in the last frame. $d<20$ m is considered near range ($~12\%$ of samples), $20$ m $\geq d > 45$ m medium range ($~65\%$ of samples) and $d \geq 45$ m far range ($~23\%$ of samples). For each of those ranges different difficulties arise in the estimation. While in near range examples the perspective on vehicles can shift drastically in between instances and over time for individual instances, for far range samples the estimation is limited by the pixel resolution of the data.

From the provided data a method should be developed, which is able to infer velocity as well as position of vehicles specified by a given bounding box in the last frame. For evaluation a test set consisting of 269 clips, or 375 vehicle tracks is provided, which is structurally identical to the training data, with the only difference being the absence of ground truth position and velocity data.

\subsection{Ablation study}

To investigate the impact of the features used in our initial approach on velocity estimation accuracy, we have conducted an ablation study to test combinations of all features and activation functions.

\paragraph{Validation split.}

In order to be able to evaluate the performance of our approach we had to first split the available training data into a training and validation set. Earlier we used 80/20 random splits, but for more robust evaluation we decided to take into account the distribution of near, medium and far range samples (12/65/23\%) and also the similarity of the test sequences. Since the training samples are sequences from a fixed set of drives, we decided to split training and validation splits such that the validation samples stem from unseen drives.

To achieve this we compute the 4096 dimensional \emph{VGGNet fc2} feature vector \cite{simonyan14}, reduce it to 2D space using \emph{t-sne} \cite{maaten08} and then cluster the resulting features into 7 clusters using k-means. We have then chosen a cluster consisting of 10\% of the total training samples with a (14/63/21\%) near, medium, far range distribution as validation set. The remaining data is used for training after defining a fixed 5-fold split for cross validation.

\paragraph{Training.}

For each of the 5 train/test splits, 10 models with fixed $3\times40$ topologies are trained according to the same paradigm as described above. Here, training was stopped early if the velocity mean squared error (\ref{MSE}) did not improve in the course of 500 iterations. The ablation was performed both training models on the whole data as well as training individual models on near, medium and far data splits.

\paragraph{Results.}

The results of the ablation study described above are shown in \tblref{tab:best_results}. It can be seen that depth and motion cues can indeed help improve results of vehicle velocity estimation, but their impact is limited by range. Both depth and optical flow estimates show significantly degrading performances for distances larger than 20m (above near range). Within near range on the other hand, where rapid changes in perspective can occur, they are superior in performance to using only tracking.

For medium range, where vehicles are predominantly viewed from their rear only, tracking cues only yield superior performance. For far range the pixel level appearance of vehicles changes so slightly at different velocities, that a robust estimation of their velocity can not be achieved by the method used.

\begin{table}[H]
\vspace{-5pt}
\centering
	\begin{tabular}{c|c|c|c|c}
	
	• & Motion & Depth & Tracking & MLP \\ 
	\hline 
	Hardware & GPU & GPU & CPU & GPU \\ 
	Timing & 344 ms & 69 ms & 7 ms & 3 ms \\ 
	\hline 
	\end{tabular} 
	\vspace{-7pt}
	\caption{\label{tab:timings}Runtime of the individual stages of our method for a single frame on Intel\textregistered  Core\textsuperscript{\tiny TM} i7-5930K, 32GB, NVIDIA Titan X. For depth and motion estimation inference times only are indicated.}
	\vspace{-10pt}
\end{table}

Using tracking cues only is also the most efficient, as shown in \tblref{tab:timings}. Inference of our model is generally very fast and thus the feature extraction is the biggest performance bottleneck of our approach. While using motion, depth and tracking cues provides a theoretical estimation rate of 2 frames per second, with tracking cues 100 frames can be estimated in one second.

\subsection{Challenge results}

\begin{figure*}[h]
	\vspace{-15pt}
	\captionsetup[subfigure]{labelformat=empty}
	\captionsetup[subfloat]{captionskip=0pt}
	\captionsetup[subfloat]{aboveskip=20pt}
	 \captionsetup[subfigure]{position=top}

	\centering
	
	\subfloat[\small  ]{\includegraphics[width=0.32\textwidth]{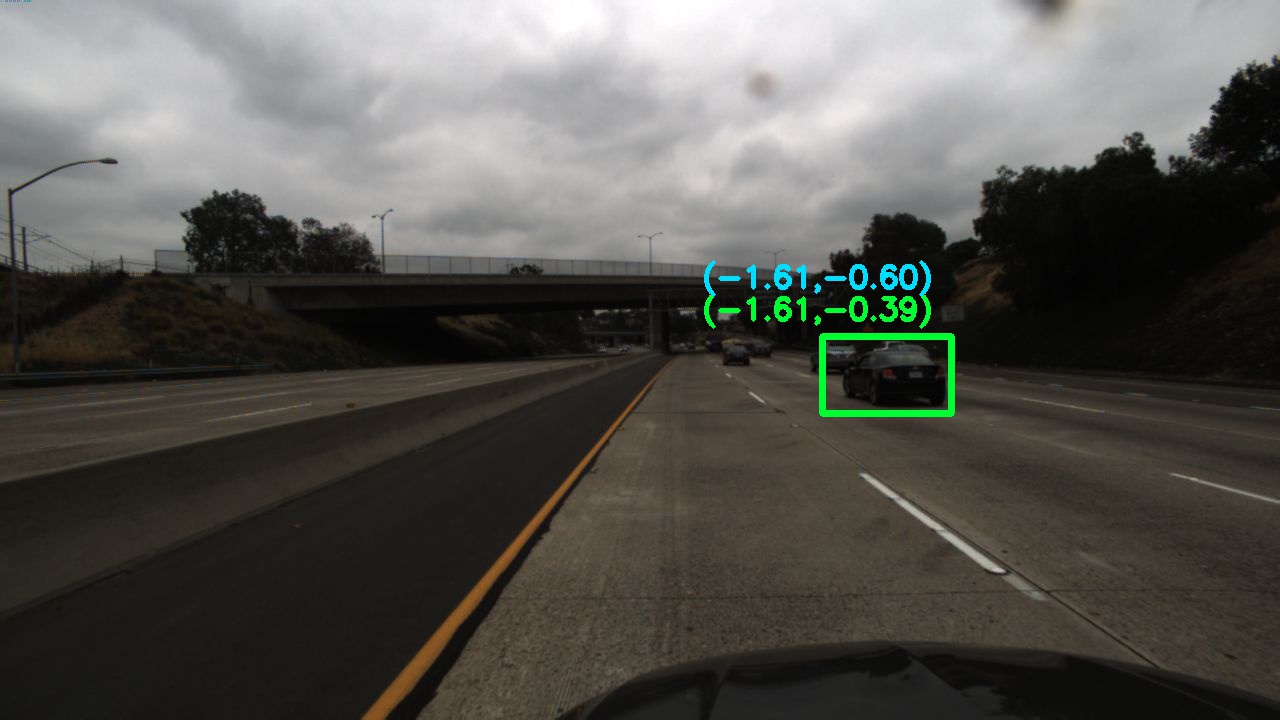}}
	\hspace{5pt}
	\subfloat[\small ]{\includegraphics[width=0.32\textwidth]{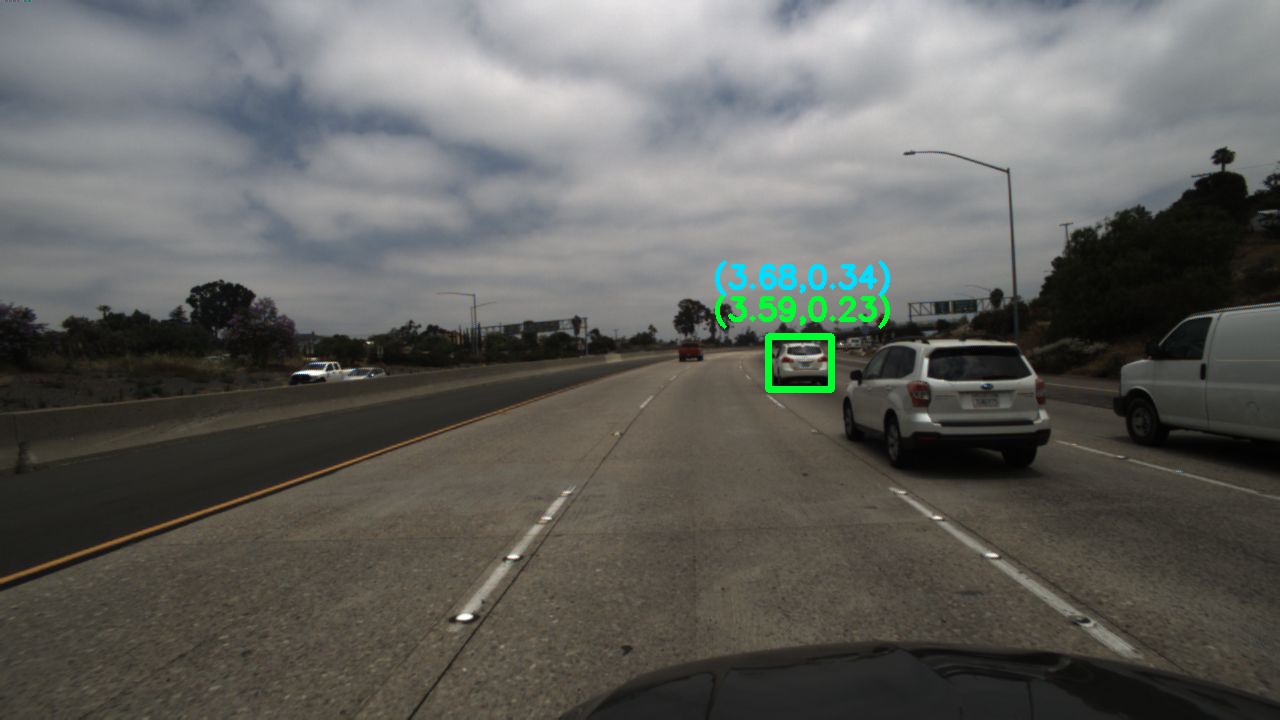}}
	\hspace{5pt}
	\subfloat[\small  ]{\includegraphics[width=0.32\textwidth]{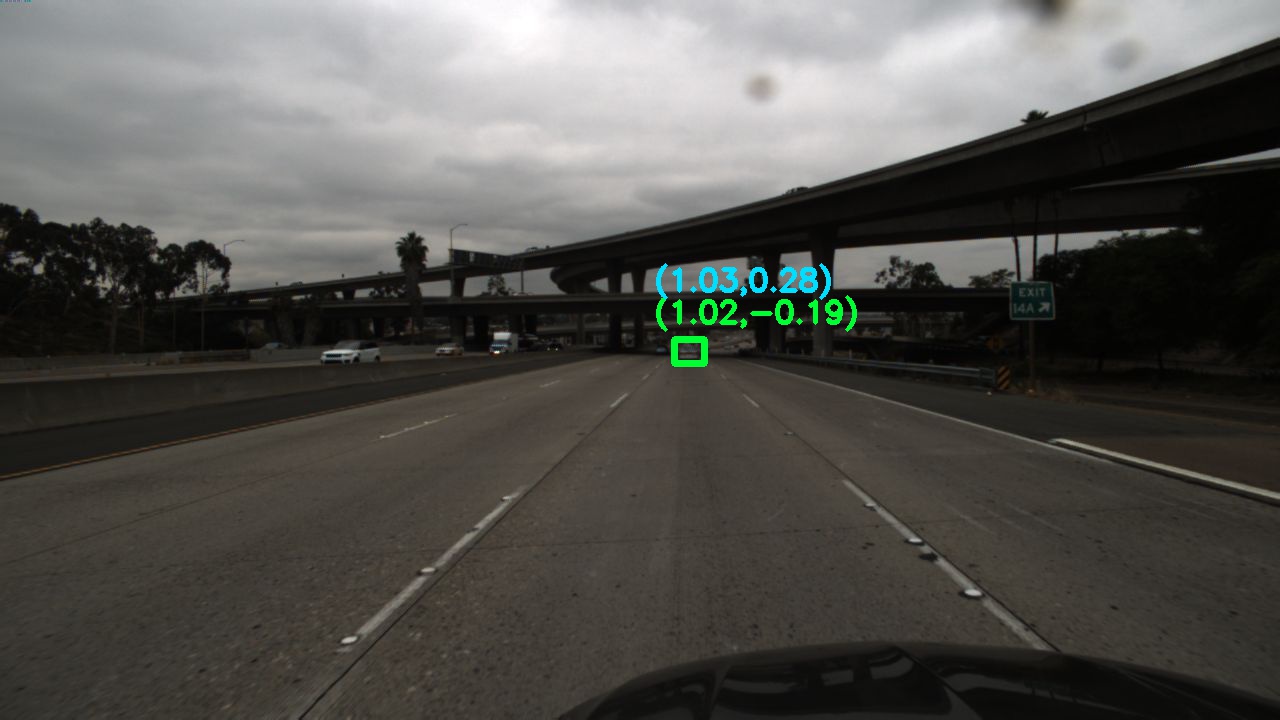}}

\vspace{-7pt}
\caption{Qualitative results on the validation data, near medium and far range (f.l.t.r.). In each example one frame of a sequence from the validation split is shown, annotated with the estimated velocity in m/s for $(x,y)$ coordinates. Blue is ground truth, green the estimated velocity.
}
\label{fig:quali_results}
\vspace{-15pt}
\end{figure*}

On the leaderboard of the CVPR2017 Autonomous Driving velocity estimation challenge, the individual entries are ranked by overall average velocity mean squared error. For each given set of samples $C$, in this case the ranges near, medium and far, the error is evaluated according to (\ref{MSE}) and then averaged (\ref{average}).

{
	\begin{gather}
	  E_{V,C} = \frac{1}{|C|} \sum_{c \in C} || V_c - \hat{V}_c ||^2  \label{MSE}\\
	  E_{V} = \frac{E_{V,near}+E_{V,med}+E_{V,far}}{3}. \label{average}
	\end{gather}
}

In correspondence to the given evaluation metric, the results achieved on the test set are depicted in \tblref{tab:test_results}. The entry \emph{\textbf{Ours} full} is the winning approach described above, using individual models trained on all three ranges separately, while incorporating tracking, flow and depth features. \emph{\textbf{Ours} tracking} are the best performing models using only tracking cues obtained in the ablation study, which are again separate models for each range, but trained on the whole data. Notably, the ground-truth accuracy is at around 0.71 m/s which is obtained from a combination of LiDAR and radar sensors (this number was communicated by the challenge organizers). Therefore, our average estimation 1.12 m/s error (corresponding to 1.25 m$^2$/s$^2$ MSE) is relatively close to a LiDAR and radar solution, but uses only videos recorded from a monocular dash cam.

\begin{table}[H]
\centering
	\begin{tabular}{l|c|c|c|c}

	• & $E_{V}$ & $E_{V,near}$ & $E_{V,med}$ & $E_{V,far}$ \\ 
	\hline 
	\textbf{Ours} tracking* & \textbf{1.25} & \textbf{0.12} & \textbf{0.54} & 3.11 \\
	\textbf{Ours} full & 1.30 & 0.18 & 0.66 & \textbf{3.07} \\ 
	 Rank2 team& 1.50 & 0.25 & 0.75 & 3.50 \\ 
	 Rank3 team& 2.90 & 0.55 & 2.21 & 5.94 \\ 
 	 Rank4 team& 3.54 & 1.46 & 2.74 & 6.42 \\
	\hline
	
	\end{tabular}
	\vspace{5pt}
	\caption{\label{tab:test_results}Challenge leaderbord top 5. \emph{\textbf{Ours} full} is the winning approach. \emph{\textbf{Ours} tracking} denotes the best performing tracking only model. {\footnotesize *submitted post deadline}}
\end{table}

\section{Conclusion}
This paper documents the winning entry at the CVPR2017 vehicle velocity estimation challenge. We have proposed a light-weight approach for directly regressing vehicle velocities from their tracks in monocular video sequences. By comparing complementary features for vehicle velocity estimation, we find that light-weight trajectory based features outperform depth and motion cues extracted from deep ConvNets. Our approach is real-time capable on a single CPU and outperforms all competing entries in the velocity estimation challenge. Future work shall address an end-to-end system for joint tracking and estimation.

\paragraph{Acknowledgments.} We are grateful for discussions with
Axel Pinz.  The GPUs used for this research were donated by NVIDIA.

{\small
	\bibliographystyle{ieee}
	\bibliography{shortstrings,references}
}

\end{document}